\documentclass[runningheads]{llncs}
\usepackage{amsmath,amssymb,amsfonts}
\usepackage{graphicx}
\usepackage[hidelinks, colorlinks=true]{hyperref}
\usepackage[linesnumbered,ruled,vlined]{algorithm2e}

\usepackage{blindtext}
\usepackage{comment}
\usepackage{dirtytalk}

\usepackage[table,xcdraw]{xcolor}
\usepackage{float}

\usepackage{subfig}

\usepackage{lipsum}  
\usepackage{multirow}

\usepackage{dirtytalk}

\usepackage{caption}
\usepackage[export]{adjustbox}

\usepackage{tabularx}

\usepackage[vlines]{tabularht}
\usepackage{pbox}
\usepackage{booktabs}
\usepackage{longtable}

\usepackage{stix,bbding,pifont,utfsym,fontawesome}
\usepackage{mathtools, cuted}

\usepackage[numbers,sort&compress, sectionbib]{natbib}

\begin{document}

\title{Motion-enhanced Cardiac Anatomy Segmentation via an Insertable Temporal Attention Module}

\titlerunning{Motion-enhanced Cardiac Anatomy Segmentation Using Novel Temporal Attention}

% \author{Md. Kamrul Hasan\inst{1}\orcidID{0000-1111-2222-3333} \and
% Guang Yang\inst{1}\orcidID{1111-2222-3333-4444} \and
% Choon Hwai Yap\inst{1}\orcidID{2222--3333-4444-5555}}

\author{Md. Kamrul Hasan \and
Guang Yang\and
Choon Hwai Yap}

\authorrunning{M. K. HASAN et al.}
% First names are abbreviated in the running head.
% If there are more than two authors, 'et al.' is used.

\institute{Department of Bioengineering, Imperial College London, UK 
\email{\{k.hasan22,g.yang,c.yap\}@imperial.ac.uk}}

\maketitle              % typeset the header of the contribution
\begin{abstract}
Cardiac anatomy segmentation is useful for clinical assessment of cardiac morphology to inform diagnosis and intervention. Deep learning (DL), especially with motion information, has improved segmentation accuracy. However, existing techniques for motion enhancement are not yet optimal, and they have high computational costs due to increased dimensionality or reduced robustness due to suboptimal approaches that use non-DL motion registration, non-attention models, or single-headed attention. They further have limited adaptability and are inconvenient for incorporation into existing networks where motion awareness is desired. Here, we propose a novel, computationally efficient Temporal Attention Module (TAM) that offers robust motion enhancement, modeled as a small, multi-headed, cross-temporal attention module. TAM’s uniqueness is that it is a lightweight, plug-and-play module that can be inserted into a broad range of segmentation networks (CNN-based, Transformer-based, or hybrid) for motion enhancement without requiring substantial changes in the network’s backbone. This feature enables high adaptability and ease of integration for enhancing both existing and future networks. Extensive experiments on multiple 2D and 3D cardiac ultrasound and MRI datasets confirm that TAM consistently improves segmentation across a range of networks while maintaining computational efficiency and improving on currently reported performance. The evidence demonstrates that it is a robust, generalizable solution for motion-awareness enhancement that is scalable (such as from 2D to 3D). The code is available at \url{https://github.com/kamruleee51/TAM}.

\end{abstract}

\section{Introduction}
\label{Introduction}
Cardiac anatomy segmentation quantifies functional parameters like chamber shape, volumes, wall thickness, ejection fraction (EF), and stroke volume \cite{bai2018automated}. Traditional methods like region growing \cite{adams1994seeded} and template matching \cite{chen2009automated} often require manual intervention and struggle with low-quality cardiac images. As a result, segmentation remains largely manual or semi-automatic, making it time-consuming and prone to imprecision. Despite clinical guidelines recommending repeated measurements over three cardiac cycles for accuracy \cite{lang2015recommendations}, this is often neglected due to time constraints, further reducing precision.
\begin{figure*}[!t]
\centering
\includegraphics[width=1.0\textwidth]{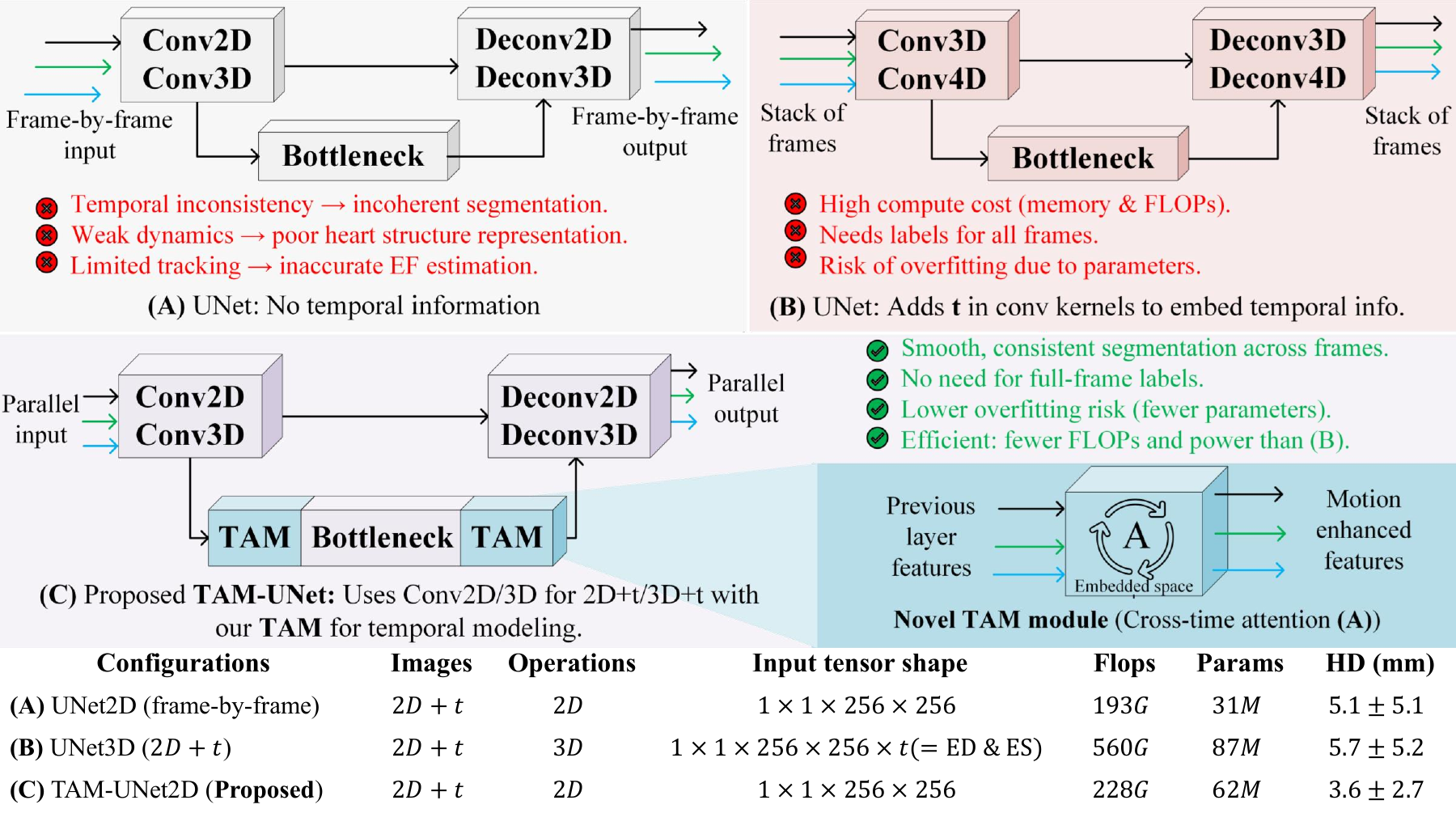}
\caption{\textbf{(A)} The baseline UNet (2D or 3D) processes each input frame independently, neglecting temporal information.
\textbf{(B)} A conventional temporal UNet integrates temporal context using Conv3D for 2D+t \cite{wei2020temporal} or Conv4D for 3D+t \cite{myronenko20204d}.
\textbf{(C)} The proposed model enhances temporal integration by combining Conv2D (for 2D+t) or Conv3D (for 3D+t) with the novel \textit{Temporal Attention Module (TAM)}.
\textbf{Bottom Table:} Computation vs. results (Hausdorff Distance) for \textbf{(A)}, \textbf{(B)}, and \textbf{(C)} using ED and ES frames for all of these.}
\label{fig:Overview}
\end{figure*}

Advances in deep learning (DL) have enabled automatic segmentation, surpassing traditional methods. Convolutional neural networks (CNNs), FCN8s \cite{long2015fully} and UNet \cite{ronneberger2015u}, have achieved significant progress in segmentation, influencing numerous subsequent works \cite{litjens2017survey}. To address CNNs' limited spatial influence, Transformers like UNetR \cite{hatamizadeh2022unetr} and SwinUNetR \cite{hatamizadeh2021swin} were introduced to enhance global feature extraction, improve transferability, and complement CNNs. In terms of including temporal information, the need for this is intuitive, as humans often find it easier to distinguish cardiac structures from non-cardiac ones in videos compared to still images. Its incorporation mitigates transient noise and signal losses and enhances the ability to identify cardiac structures \cite{chan2021full}. A common approach for such temporal awareness involves adding a time dimension and using higher-dimensional models, such as 3D networks for 2D+t images \cite{wei2020temporal} or 4D networks for 3D+t \cite{myronenko20204d}. However, these methods face challenges like the need for extensive manual annotations, high computational costs (especially for 4D networks), and increased risk of overfitting on small datasets (Fig.~\ref{fig:Overview}B). Hybrid approaches, which combine segmentation with temporal registration \cite{xue2022improved, wei2020temporal}, Conv-LSTM \cite{li2019recurrent}, or two-stage pipelines \cite{wei2023co}, have shown promise. However, they remain computationally intensive and rely heavily on ground truth segmentations. Further, the temporal registration algorithms used are often imperfect. Finally, to date, motion-informed DL networks, including those using temporal attention \cite{wu2022semi, ahn2021multi}, are standalone frameworks with limited adaptability for integration into other networks or a future network.

To address these limitations, we propose the novel temporal attention module (TAM), a lightweight module that can seamlessly be integrated into CNN, Transformer, and hybrid networks to incorporate motion awareness. Unlike prior methods, it uses a few labeled frames and propagates information through TAM attention to leverage unlabeled frames, reducing the need for all frame manual annotations. Our key contributions are: (1) \textbf{Novel TAM:} A multi-headed, cross-time attention mechanism based on KQV projection, modeled as a small plug-in for existing networks to enable effective capture of long-range cardiac dynamic changes while minimizing computational overhead. (2) \textbf{Varied Networks' Adaptability:} A plug-and-play module that can be conveniently integrated into various network backbones to enhance their motion awareness, such as UNet, FCN8s, UNetR, SwinUNetR, I$^2$UNet \cite{dai2024i2u}, and DT-VNet \cite{cai2024dt}, and (3) \textbf{Generalizable, Scalable, and Low Cost:} TAM is generalizable to various datasets with varying modalities and quality: Ultrasound (2D CAMUS \cite{leclerc2019deep} and 3D MITEA \cite{zhao2023mitea}) and 3D MRI (ACDC \cite{bernard2018deep}). It is scalable from 2D to 3D and improves performance while maintaining low computational overhead compared to the conventional alternative in Fig.~\ref{fig:Overview} (B).

\begin{figure*}[!b]
\centering
\includegraphics[width=1.0\textwidth]{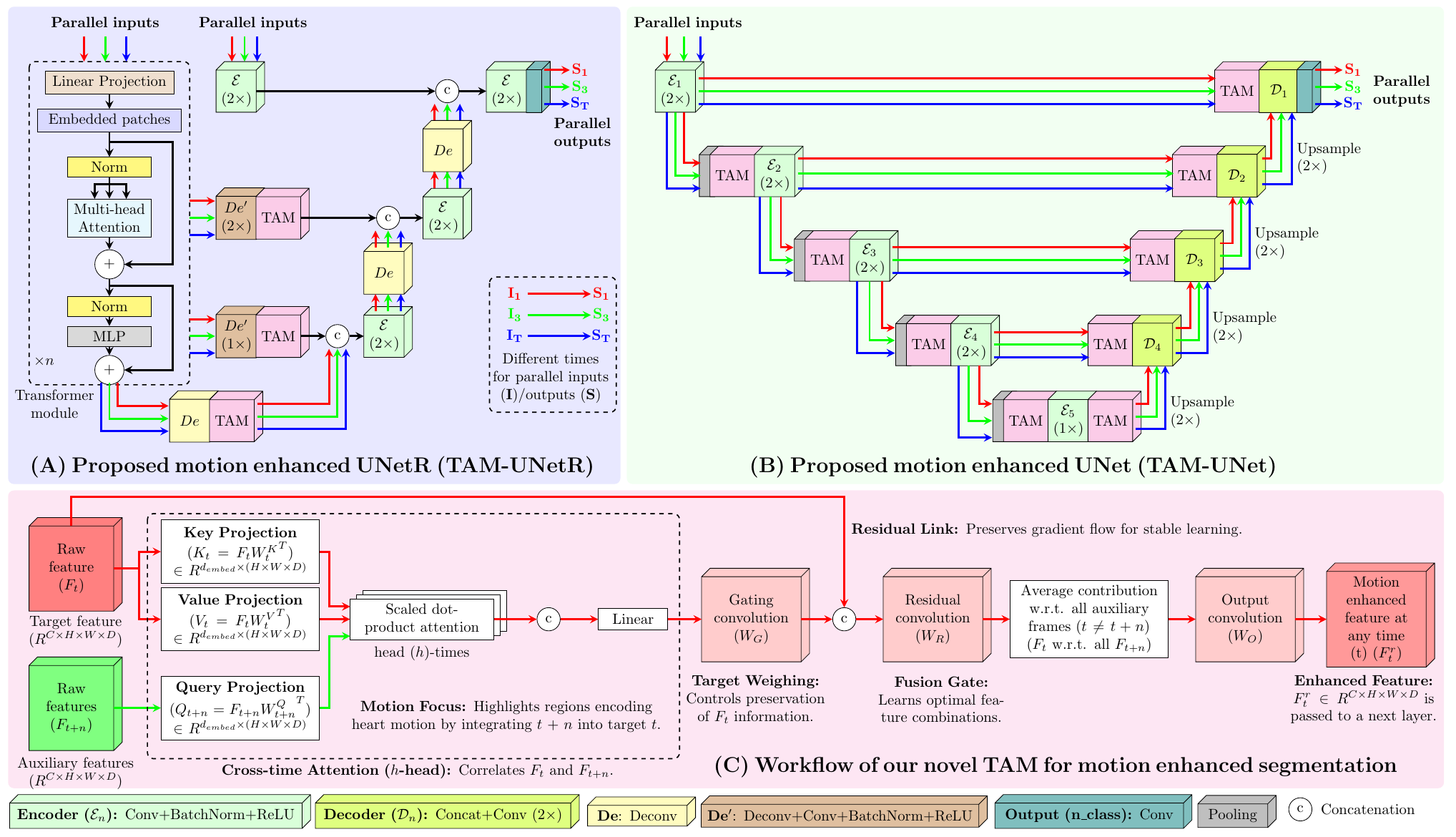}
\caption{In \textbf{(A-B)}, TAM is added to the transformer-based UNetR and the CNN-based UNet. As illustrated in \textbf{(C)}, TAM uses multi-headed cross-time attention for cross-frame motion refinement, enhancing segmentation quality, unlike the works in \cite{wei2020temporal, myronenko20204d} (Fig.~\ref{fig:Overview}).} 
\label{fig:methods}
\end{figure*}

\section{Methodology and Materials}
\label{Methodology}
\subsection{Proposed motion-enhanced segmentation}
Given a cardiac dataset, denoted as $D = {\big(V^n, S^n\big)}_{n=1}^N$, where $V^n \in \mathbb{R}^{T \times H \times W}$ (for 2D) or $V^n \in \mathbb{R}^{T \times H \times W \times D}$ (for 3D) represents a temporal sequence of $T$ frames. Each frame has a spatial resolution of $H \times W$ for 2D or $H \times W \times D$ for 3D. $S^n$ denotes the anatomical segmentation mask for $V^n$. Our objective is to leverage the temporal dimension ($T$) efficiently and robustly for smooth and consistent segmentation across frames. We achieve this via the TAM, a plug-and-play module that integrates seamlessly into various backbone segmentation algorithms. As shown in Fig.~\ref{fig:methods}, TAM explicitly models temporal relationships, focusing on relevant frames and regions to segment the target frame by enhancing temporal coherence, refining motion-sensitive areas, and suppressing noise.

\begin{algorithm*}[!t]
\KwIn{Feature maps $\{F_1, F_2, \dots, F_T\} \in \mathbb{R}^{C \times H \times W \times D}$, $C$: channel numbers}

\KwOut{Motion-enhanced feature maps $\{F_1^{\text{r}}, F_2^{\text{r}}, \dots, F_T^{\text{r}}\} \in \mathbb{R}^{C \times H \times W \times D}$}

Compute query, key, and value projections and split them into multiple heads $H_{\text{heads}}$:
\[Q_t = F_t W_q^T+b_q, \, K_t = F_t W_k^T+b_k, \, \text{and}\, V_t = F_t W_v^T+b_v,\, \text{where}\, Q_t, K_t, V_t \in \mathbb{R}^{d_{\text{embed}} \times (H \times W \times D)}\]
\[Q_t^h, K_t^h, V_t^h \in \mathbb{R}^{\frac{d_{\text{embed}}}{H_{\text{heads}}} \times (H \times W \times D)}, \quad \forall h \in \{1, 2, \dots, H_{\text{heads}}\}\]

Compute attention for the $i^{th}$ time with respect to $j^{th}$ time ($i\neq j$) across all heads $h$:
    \[
    A_{i \leftarrow j}^h = \text{softmax}\left( \frac{Q_i^h K_j^{h\top}}{\sqrt{d_{\text{embed}}/H_{\text{heads}}}} \right) V_j^h,\quad A_{i \leftarrow j}^{\text{multi-head}} = \text{Concat}(A_{i \leftarrow j}^1, A_{i \leftarrow j}^2, \dots, A_{i \leftarrow j}^{H_{\text{heads}}}) 
    \]
    
Apply a gating operation ($W_G$) and residual concatenation operation ($W_{R}$):
\[
G_{i \leftarrow j} = W_G * A_{i \leftarrow j}^{\text{multi-head}}, \quad A_{i \leftarrow j}^{\text{gated}} = A_{i \leftarrow j}^{\text{multi-head}} \odot G_{i \leftarrow j}
\]
\[
F_i^{\text{combined}} = \text{Concat}(F_i, A_{i \leftarrow j}^{\text{gated}}), \quad F_i^{R} = W_{R} * F_i^{\text{combined}}
\]

Attention aggregation of all contributing frames and apply output operation ($W_{O}$):
\[
F_i^{\text{Avg}} = \frac{1}{T-1} \sum_{j \neq i}^T F_i^{R}, \quad F_i^{\text{r}} = W_{\text{O}} * F_i^{\text{Avg}}
\]

Return refined motion-enhanced feature maps $\{F_1^{\text{r}}, F_2^{\text{r}}, \dots, F_T^{\text{r}}\} \in \mathbb{R}^{C \times H \times W \times D}$

\caption{Multi-headed cross-time attention algorithm of our novel TAM.}
\label{algo:cross_attention}
\end{algorithm*}

\textbf{Integrating TAM into a backbone segmentation model:}
Fig.~\ref{fig:methods} shows how two well-known networks can be coupled with our proposed TAM module (Fig.~\ref{fig:methods}C). Segmentation networks using UNetR and UNet that initially lack motion awareness are enhanced by integrating TAM at various locations in the network (Fig.~\ref{fig:methods} (A, B)), forming TAM-UNetR and TAM-UNet. Other study networks are similarly modified.

To enable TAM insertion, the networks are slightly modified to input multiple frames (e.g., 2-5) in parallel for inter-frame motion information propagation. For each input frame ($V_t^n,t\in T$), the network layer generates a feature map $F_t$, which is enhanced by the TAM module to produce motion-refined $F_t^\text{r}$, for the next layer. The location(s) at which TAM modules are placed can be strategically optimized to be at specific encoder or decoder layers of the backbone network through ablation studies (Section~\ref{Results}) for improved performance. The addition of TAM adds minimal computation costs, as the same module with shared weights is used at different layers of the network. Compared to adding the time dimension to the input tensor to achieve motion awareness as in \cite{wei2020temporal, myronenko20204d}, TAM’s approach is much more efficient (shown in Table~\ref{tab:TAM_ablation}).

\subsubsection{Temporal attention module (TAM):}
As shown in Fig.~\ref{fig:methods}C, TAM is integrated between two layers of the model. TAM receives $T$ number of temporal feature maps $\{F_{t}\} \in \mathbb{R}^{C \times H \times W \times D}$ ($t\in T$) from the previous layer and facilitates pairwise temporal feature interaction (between $F_{t=i}$ and $F_{t=j}$, where $i\neq j$) in the embedded space to refine the input features and pass refined features $\{F_{t}^{\text{r}}\} \in \mathbb{R}^{C \times H \times W \times D}$ ($t\in T$) to the subsequent layer. This identifies key cardiac regions and frames for a refined representation. Algorithm~\ref{algo:cross_attention} outlines the TAM processing steps, which are explained below. 

\textbf{Step-1:} Features are transformed with trainable weights ($W_q$, $W_k$, $W_v$) and biases ($b_q$, $b_k$, $b_v$) to generate the Query ($Q_t$), Key ($K_t$), and Value ($V_t$). $K$ and $V$ encode the target frame, while $Q$ encodes the alternate frame. The embeddings are split into $H_\text{heads}$ subspaces to capture diverse temporal patterns and spatial-temporal associations.
\textbf{Step-2:} Cross-time attention ($A^h$) is computed between the target frame $i$ and all other frames $j \neq i$ and is scaled by $\sqrt{d_\text{embed}/H_\text{heads}}$ to stabilize gradients and prevent large dot product values. Softmax is used to normalize attention scores to prioritize relevant spatial-temporal areas.
$A^h$ is then concatenated to combine diverse attention outputs. It maintains the embedding dimension $d_\text{embed}$ for downstream compatibility while capturing crucial motion refinement.
\textbf{Step-3:} 
To handle appearance variations and noise, frames are weighted differently \cite{lu2019see} using a self-gating mechanism that assigns a confidence score to relevant areas. This involves a convolution ($W_G$), bias ($b_G$), and activation ($\sigma$) to control each region's contribution. The gate adjusts the reference frame's influence, and attention summaries are updated with Hadamard products ($\odot$) between $A_{i \leftarrow j}^{\text{multi-head}}$ and $G_{i \leftarrow j}$, forming a gated cross-attention. The concatenation of original feature maps with the gated attention output forms $F^{\text{combined}}$, enhancing temporal affinities while preserving spatial structure. $W_R$ aggregates local spatial details, batch normalization stabilizes training, and ReLU after $W_R$ introduces non-linearity to model complex spatial-temporal patterns.
\textbf{Step-4:} 
Motion features across all frames $j \neq i$ are aggregated, with a convolution reducing dimensionality while preserving motion-relevant details.

\subsection{Datasets and Implementation}
\label{Datasets}
The \textbf{CAMUS 2D echo} \cite{leclerc2019deep} includes 500 patients with apical 2-chamber and 4-chamber views (details in \cite{leclerc2019deep}). 50 are selected for testing, while the remaining are used for training and validation.
The \textbf{MITEA 3D echo} \cite{zhao2023mitea} includes 134 patients with 268 samples at end-diastole and end-systole. The training-validation/testing distribution: 328 (304/24) healthy, 56 (48/8) with left ventricular hypertrophy, 48 (36/12) with cardiac amyloidosis, 40 (32/8) with aortic regurgitation, 32 (28/4) with hypertrophic cardiomyopathy, 24 (20/4) with dilated cardiomyopathy, and 8 (4/4) with heart transplant recipients.
The \textbf{ACDC 3D MRI} \cite{bernard2018deep} consists of cardiac MRI scans from 100 patients with detailed segmentation annotations. It is divided into 70/10/20 for training, validation, and testing.

\textbf{\textit{Experimental Setup:}} We use the Dice Similarity Coefficient (DSC) for overlapping precision, Hausdorff Distance (HD) for border irregularity, and Mean Average Surface Distance (MASD) to assess segmentation boundary accuracy. Experiments were run in \texttt{PyTorch} on Ubuntu with 4$\times$3090\,Ti GPUs, using Adam (LR of $10^{-4}$) for 250 epochs; CAMUS (2D, batch=8, $256^2$), MITEA \& ACDC (3D, batch=4, $128^3$ / $160\!\times\!160\!\times\!64$).

\textbf{\textit{Loss function:}} In alignment with recent literature \cite{shaker2024unetr++}, the loss function ($\mathcal{L}$) is formulated as follows using the true and predicted anatomical masks ($S_T$ and $S_P$) for all $N$ elements (pixel/voxel). 
\[
\mathcal{L}(S_T, S_P) = \frac{1}{\mathcal{C}}\sum_{i\in \mathcal{C}} \Biggl(1- \frac{2\cdot|S^i_T \cap S^i_P|}{|S^i_T| + |S^i_P|}+ \frac{1}{N}\sum_{x \in N} S^i_T (x) \log S^i_P (x)  \Biggl), \,\mathcal{C} \text{: class numbers.}
\]

\section{Results and Discussion}
\label{Results}
\begin{table*}[!b]
\caption{Results on CAMUS for TAM-UNet2D (8 heads) and two baselines on ED-ES.}
\label{tab:TAM_ablation}
\centering
\fontsize{8pt}{8pt}\selectfont
\begin{tabular}{lcccccc}
\toprule
\textbf{Configurations (C)} & & \textbf{DSC ($\uparrow$)} & \textbf{HD ($\downarrow$)} & \textbf{MASD ($\downarrow$)} & \textbf{FLOPs ($\downarrow$)} & \textbf{Params ($\downarrow$)} \\
\midrule
UNet2D (without motion) & \textbf{C1}   & $0.913$ & $5.11\,mm$ & $1.13\,mm$ & $193\,G$ & $31\,M$ \\
UNet3D (2D+t) (with motion) & \textbf{C2}   & $0.915$ & $5.65\,mm$ & $1.15\,mm$ &  $560\,G$ & $87\,M$ \\\midrule
\multicolumn{7}{c}{Different configurations of our motion-enhanced TAM-UNet2D (Fig.~\ref{fig:methods}B)} \\ \midrule
TAM with $\mathcal{E}_5$ only & \textbf{C3}  & $0.921$ & $3.90\,mm$ & $0.986\,mm$ &  $221\,G$ & $58\,M$ \\

TAM with $\mathcal{E}_4\,\&\,\mathcal{E}_5$ & \textbf{C4}  & $0.922$ & $3.63\,mm$ & $0.961\,mm$ &  $228\,G$ & $62\,M$\\

TAM with $\mathcal{E}_3,\,\mathcal{E}_4\,\&\,\mathcal{E}_5$ & \textbf{C5}   & $0.923$ & $3.51\,mm$ & $0.954\,mm$ &  $237\,G$ & $66\,M$ \\

TAM with $\mathcal{E}_5$  and $\mathcal{D}_4$ & \textbf{C6}  & $0.923$ & $3.68\,mm$ & $0.952\,mm$ &  $253\,G$ & $65\,M$ \\

TAM with $\mathcal{E}_5$ and $\mathcal{D}_3\,\&\,\mathcal{D}_4$ & \textbf{C7}   & $0.923$ & $4.05\,mm$ & $0.966\,mm$ &  $316\,G$ & $66\,M$ \\

TAM with $\mathcal{E}_4\,\&\,\mathcal{E}_5$ and $\mathcal{D}_4$ & \textbf{C8}  & $0.923$ & $3.55\,mm$ & $0.957\,mm$  & $270\,G$ & $66\,M$ \\

TAM with $\mathcal{E}_4\,\&\,\mathcal{E}_5$ and $\mathcal{D}_3\,\&\,\mathcal{D}_4$ & \textbf{C9}  & $0.921$ & $4.04\,mm$ & $0.980\,mm$ &  $332\,G$ & $68\,M$\\

TAM with $\mathcal{E}_3,\,\mathcal{E}_4\,\&\,\mathcal{E}_5$ and $\mathcal{D}_4$ & \textbf{C10}  & $0.922$ & $3.79\,mm$ & $0.969\,mm$  & $260\,G$ & $67\,M$ \\

TAM with $\mathcal{E}_3,\,\mathcal{E}_4\,\&\,\mathcal{E}_5$ and  $\mathcal{D}_3\,\&\,\mathcal{D}_4$ & \textbf{C11}  & $0.920$ & $3.80\,mm$ & $0.977\,mm$ & $323\,G$ & $68\,M$\\
\bottomrule
\end{tabular}
\end{table*}
We optimized the multi-headed TAM design by fine-tuning the number of attention heads, input time frames, and the locations where TAM is inserted in the network to balance performance and computational efficiency. An ablation study evaluated various TAM configurations in the UNet and compared them with baseline UNet2D and UNet3D (2D+t) models. Results in Table~\ref{tab:TAM_ablation} show that various configurations of TAM-UNet2D with two input frames (ED, ES) significantly improve DSC, HD, and MASD ($p < 0.05$) compared to the baseline models while remaining efficient. For an $L$-layer network with $k \times k$ kernels and $C$ channels, the FLOPs are proportional to $L k^2 C^2HW$ for UNet2D, $L k^3 C^2HW$ for UNet3D, and $L k^2 C^2 HW + T^2C^2 HW$ for our proposed TAM-UNet. As \( k^3 \gg k^2 + T^2 \), UNet3D is much more computationally expensive for small \( T \). With only ED and ES frames (\( T = 2 \)), TAM-UNet significantly improves segmentation quality. Including all frames may improve UNet3D's results \cite{wei2020temporal} but adds high computational overhead and also requires all the frames to be manually segmented. On the other hand, in our TAM-UNet, to segment any frame, the network incorporates the motion information from other \(T-1\) frames' embedding through the TAM's attention, not necessitating those \(T-1\) frames to be manually segmented.

\begin{figure*}[!t]
   \centering
   \subfloat{\includegraphics[width=0.33\textwidth]{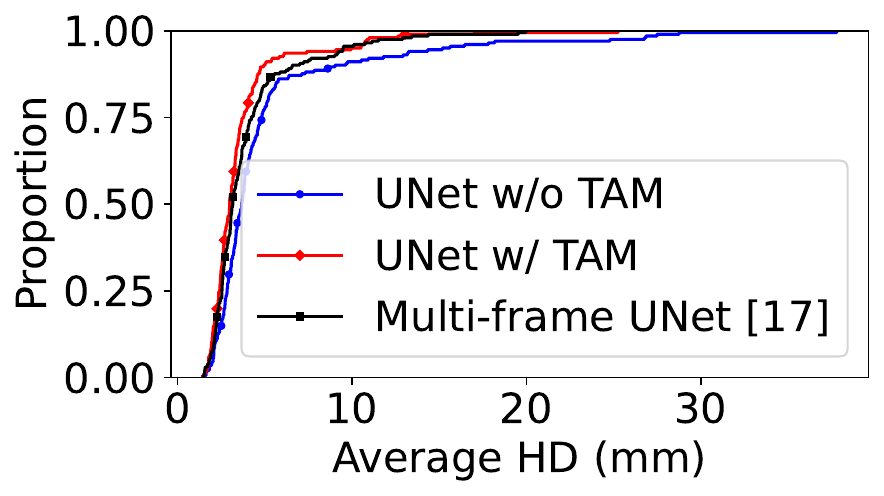}}
   \subfloat{\includegraphics[width=0.33\textwidth]{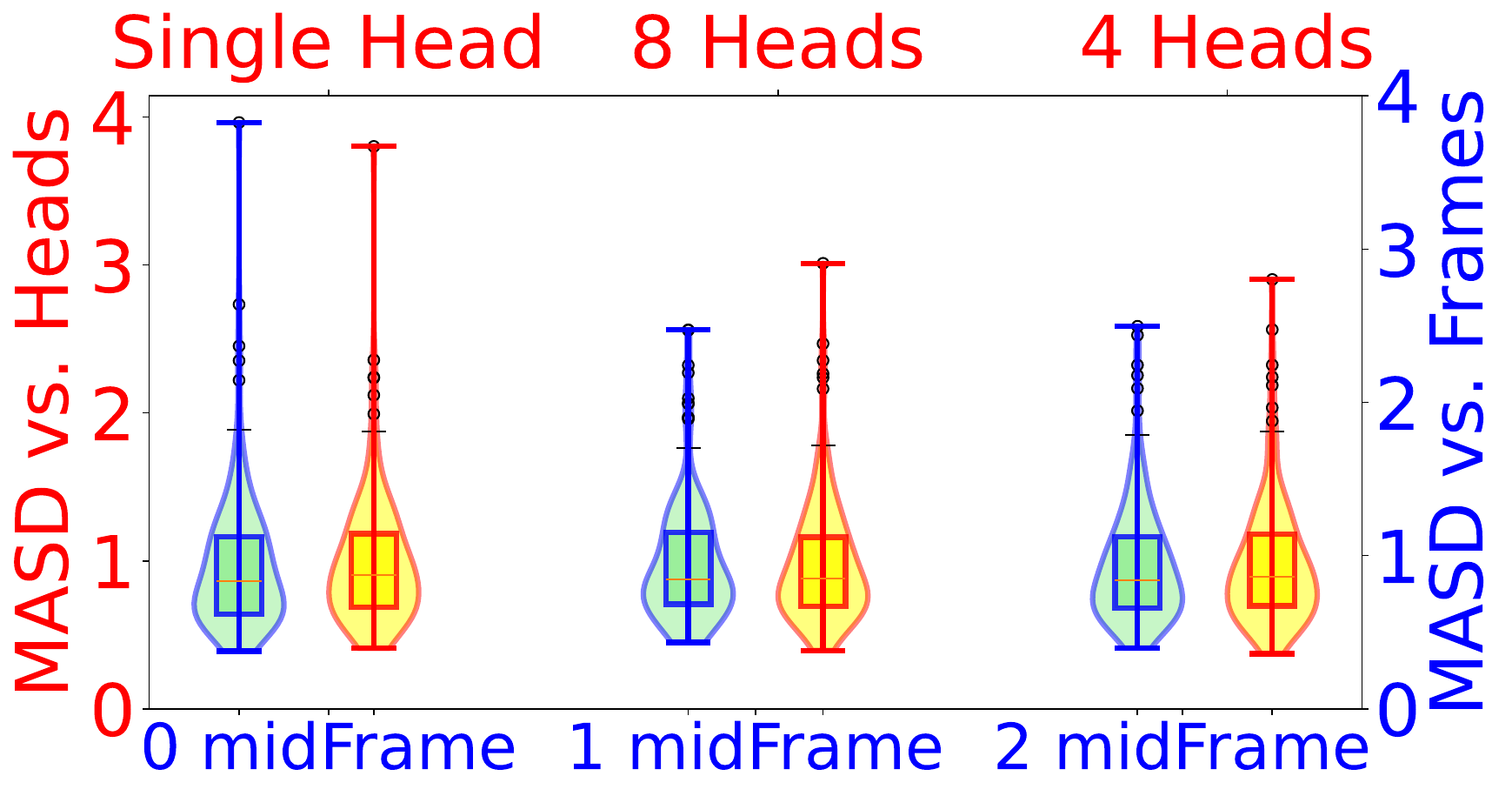}}
   \subfloat{\includegraphics[width=0.33\textwidth]{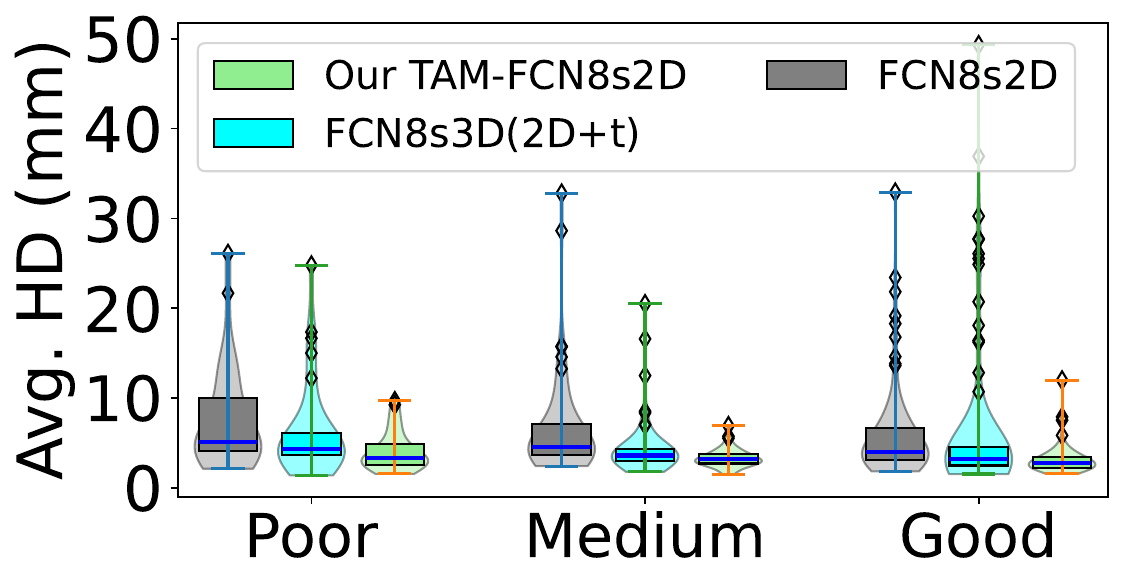}}
   \caption{\textbf{Left:} Empirical CDFs (leftward-upper shift for better accuracy) (UNet w/o \textbf{vs.} w/ TAM: \textbf{p<0.05} and  UNet w/ TAM \textbf{vs.} multi-frame UNet \cite{ahn2021multi}: \textbf{p<0.05}), \textbf{Middle:} MASD \textbf{vs.} Heads/midFrame numbers in between ED and ES, \textbf{Right:} Across image qualities.}
   \label{fig:ECD}
\end{figure*}

Comparisons between the various configurations (C3–C11) in Table~\ref{tab:TAM_ablation} demonstrate that although the location of TAM insertion impacts FLOPs and performance, results remain fairly consistent across configurations. We consider C3 and C4 to be optimal, as they balance performance and efficiency, have fewer outliers, are less complex with lower FLOPS, and use them for further testing. Ablation results with C3 and C4 show that the multi-head TAM outperforms the single-head TAM (Fig.~\ref{fig:ECD} \textbf{(middle)}), likely due to its ability to capture more diverse image features and that the 8-headed C4 TAM-UNet is the optimal setup. Additionally, including a mid-frame between ED and ES improves results compared to using only ED and ES frames, as the intermediate frames help the network bridge over the large motion between ED and ES images (Fig.~\ref{fig:ECD} \textbf{(middle)}).

We also compared our multi-head KQV design in TAM-UNet (Algorithm~\ref{algo:cross_attention}) with other multi-frame UNet models \cite{ahn2021multi, lu2019see} that use feature multiplication for attention. Using the C4 configuration and the same hyperparameters, we find that our TAM outperforms multi-frame (Fig.~\ref{fig:ECD} \textbf{(left)} and Table~\ref{tab:camus}), with a higher proportion of cases having lower HD, where other scores remain similar or better. These results demonstrate that our motion modeling design provides a clear improvement over prior approaches.

\begin{table}[!t]
    \centering
    \begin{minipage}{0.45\textwidth}
    \fontsize{8pt}{8pt}\selectfont
        \centering
    \caption{Results on \textcolor{red}{2D Echo} \cite{leclerc2019deep}}
    \label{tab:camus}
\begin{tabular}{l|ccc}
\toprule
                 Methods              & $LV_{DSC}$ ($\uparrow$)   & $LV_{HD}$ ($\downarrow$)   & $LV_{MASD}$ ($\downarrow$)       \\ \midrule 
               
                UNet \cite{ronneberger2015u}         &  $0.928$  &  $4.21\,mm$   &  $1.06\,mm$  \\
                \textcolor{blue}{\textbf{TAM}}-UNet                   &  \textcolor{blue}{$\mathbf{0.938}$}  &  $3.07\,mm$   &  \textcolor{blue}{$\mathbf{0.894}\,mm$}  \\ \midrule

                FCN8s \cite{long2015fully}                  &  $0.913$  &  $5.44\,mm$   &  $1.27\,mm$ \\
                \textcolor{blue}{\textbf{TAM}}-FCN8s                    &  $0.935$  &  $3.04\,mm$   &  $0.949\,mm$   \\ \midrule

                SwinUNetR \cite{hatamizadeh2021swin}             &  $0.908$  &  $5.60\,mm$   &  $1.42\,mm$   \\
                \textcolor{blue}{\textbf{TAM}}-SwinUNetR                    &  $0.925$  &  $4.25\,mm$   &  $1.14\,mm$   \\ \midrule                

                I$^2$UNet \cite{dai2024i2u}                    &  $0.933$  &  $3.49\,mm$   &  $1.02\,mm$ \\

                \textcolor{blue}{\textbf{TAM}}-I$^2$UNet                   &  $0.933$  &  \textcolor{blue}{$\mathbf{3.02}\,mm$}   &  $0.972\,mm$ \\\midrule
                
                SOCOF \cite{xue2022improved}                    &  $0.932$  &  $3.21\,mm$   &  $1.40\,mm$ \\
                
                CLAS \cite{wei2020temporal}                    &  $0.935$  &  $4.60\,mm$   &  $1.40\,mm$ \\
                
                Multi-frame \cite{ahn2021multi}                    &  $0.935$  &  $3.32\,mm$   &  $0.939\,mm$ \\ \bottomrule

\end{tabular}
    \end{minipage}%
    \hspace{0.9cm}
    \begin{minipage}{0.45\textwidth}
        \centering
        \fontsize{8pt}{8pt}\selectfont
        \begin{minipage}{\textwidth}
            \centering
    \caption{Results on \textcolor{red}{3D Echo} \cite{zhao2023mitea}}
    \label{tab:mitea}
\begin{tabular}{l|ccc}
\toprule
                 Methods              & $LV_{DSC}$ ($\uparrow$)   & $LV_{HD}$ ($\downarrow$)   & $LV_{MASD}$ ($\downarrow$)       \\ \midrule 
               
                UNet \cite{ronneberger2015u}         &  $0.842$  &  $9.89\,mm$   &  $2.07\,mm$  \\
                \textcolor{blue}{\textbf{TAM}}-UNet                   &  \textcolor{blue}{$\mathbf{0.853}$}  &  \textcolor{blue}{$\mathbf{9.13}\,mm$}   &  \textcolor{blue}{$\mathbf{1.90}\,mm$}  \\ \midrule

                UNetR \cite{hatamizadeh2022unetr}             &  $0.819$  &  $11.85\,mm$   &  $2.41\,mm$   \\
                \textcolor{blue}{\textbf{TAM}}-UNetR                    &  $0.830$  &  $9.55\,mm$   &  $2.16\,mm$    \\ \bottomrule
\end{tabular}
        \end{minipage}
        \begin{minipage}{\textwidth}
        \vspace{-0.10cm}
        \fontsize{7.5pt}{7.5pt}\selectfont
            \centering
    \caption{Results on \textcolor{red}{3D MRI} \cite{bernard2018deep}}
    \label{tab:acdc}
\begin{tabular}{l|ccc}
\toprule
                 Methods              & $LV_{DSC}$ ($\uparrow$)   & $LV_{HD}$ ($\downarrow$)   & $LV_{MASD}$ ($\downarrow$)       \\ \midrule 
               
                UNet \cite{ronneberger2015u}         &  $0.939$  &  $4.61\,mm$   &  $0.526\,mm$  \\
                \textcolor{blue}{\textbf{TAM}}-UNet                   &   \textcolor{blue}{$\mathbf{0.950}$}  &  $4.00\,mm$   &  \textcolor{blue}{$\mathbf{0.436}\,mm$}   \\ \midrule

                DT-VNet \cite{cai2024dt}            &   $0.926$  &  $3.86\,mm$   &  $0.692\,mm$   \\
                \textcolor{blue}{\textbf{TAM}}-DT-VNet                    &  $0.930$  &  \textcolor{blue}{$\mathbf{3.61}\,mm$}   &  $0.652\,mm$  \\ \bottomrule
\end{tabular}
        \end{minipage}
    \end{minipage}
\end{table}

\textbf{TAM integration to CNN, Transformer, and Hybrid models:}
We examine the impact of integrating our TAM module into various backbone networks for 2D echo segmentation, including CNN-based (UNet, FCN8s), Transformer-based (UNetR), and very recent hybrid (I$^2$UNet \cite{dai2024i2u}) models. Results in Table~\ref{tab:camus} show TAM can improve segmentation performance. There were modest gains for DSC and none for I$^2$UNet. However, TAM significantly ($p < 0.05$) reduces HD and MASD, suggesting an improved LV and MYO boundary precision and a reduction in erroneous segmentation islands. \textit{Supplementary video} further demonstrates a temporally consistent sequence segmentation across different time frames.

To test if TAM’s enhancement applies only to images of specific quality, we evaluated TAM’s integration with FCN8s on the CAMUS 2D echo dataset, dividing our tests into those with good, medium, and poor-quality images. Results in Fig.~\ref{fig:ECD} \textbf{(right)} show that TAM consistently improves on FCN8s3D (2D+time) and FCN8s2D across all image quality categories, suggesting a genericity of TAM across image quality. Here, the better performance of FCN8s3D over the non-motion-informed 2D version is consistent with prevailing literature that motion information improves segmentation results.

\textbf{TAM's scalability and cross-dataset genericity:}
We further tested TAM’s scalability from 2D to 3D by testing its integration with the CNN-based UNet and the Transformer-based UNetR on the MITEA 3D echo and ACDC 3D MRI datasets. Results are shown in Table~\ref{tab:mitea} and Table~\ref{tab:acdc}, showing that TAM consistently improves results ($p < 0.05$) across all metrics. As before, the most significant gains are in HD and MASD than DSC. These results show that TAM is scalable from 2D to 3D and can successfully enhance motion awareness even with increased image dimensionality. Secondly, they provide evidence that TAM has robustness across different datasets and image types, which suggests adaptability across images with different textures, image quality, and cardiac views, making it a generalizable solution. Finally, these results further reinforce the evidence for the TAM’s cross-backbone adaptability and versatility across networks.

\textbf{Comparison to SOTA:}
We compared TAM to various SOTA on the CAMUS dataset. I$^2$UNet is arguably a SOTA that promotes historical information reuse through dual paths. As discussed, our tests show that TAM can improve HD and MASD when added to I$^2$UNet ($p < 0.05$). Compared to SOTA motion-aware methods like SOCOF \cite{xue2022improved} and CLAS \cite{wei2020temporal} (Table~\ref{tab:camus}), the TAM-enhanced network achieves a similar DSC but improves on HD and MASD ($p < 0.05$). CLAS combines DL registration and segmentation and requires the input of the entire time sequence, while TAM only requires 3 time frames, has many fewer trainable parameters, and is thus computationally lighter. SOCOF relies on constraints based on optical flow for motion awareness. Although it can theoretically handle a lower number of input frames, its implementation utilizes the entire time sequence, which is likely needed to achieve the presented results. Thus, here as well, TAM is likely to be much more computationally efficient. We further evaluated the temporal attention approach by Ahn et al. \cite{ahn2021multi}, utilizing 5 frames instead of just 3 to enhance the results. Our results again show that TAM-enhanced networks have better HD and MASD but similar DSC despite needing fewer input frames. Finally, it is noteworthy that Table~\ref{tab:camus} shows that TAM may be all we need for motion enhancement, as TAM’s addition to very simple networks (FCN and UNet) can surpass complex SOTAs above.

\section{Conclusion}
\label{Conclusion}
Our proposed TAM module effectively introduces motion awareness into segmentation networks, offering improvements, particularly in HD and MASD, thus improving boundary delineation and mitigating noise and signal losses. Its plug-and-play design allows easy integration into CNN, transformer, and hybrid networks without adding significant computational overhead, enabling flexible and efficient real-world applications. TAM’s multi-head cross-time attention, combined with spatial gating, is key to its performance gains, offering a robust solution for motion-aware segmentation.

% \subsubsection*{Acknowledgements.}
% M. K. Hasan was supported by the EPSRC-DTP studentship funds (2022–2026) from the Bioengineering department of Imperial College London.
% G. Yang was supported in part by the ERC IMI (101005122), the H2020 (952172), the MRC (MC/PC/21013), the Royal Society (IEC/NSFC/211235), the NVIDIA Academic Hardware Grant Program, the SABER project supported by Boehringer Ingelheim Ltd, and the UKRI Future Leaders Fellowship (MR/V023799/1).

\subsubsection*{Disclosure of Interests.} The authors have no competing interests to declare that are relevant to the content of this article.

\renewcommand{\bibname}{References}
\bibliographystyle{splncs04nat}
\bibliography{Reference.bib}

\begin{thebibliography}{24}
\expandafter\ifx\csname natexlab\endcsname\relax\def\natexlab#1{#1}\fi
\providecommand{\bibinfo}[2]{#2}
\ifx\xfnm\relax \def\xfnm[#1]{\unskip,\space#1}\fi
%Type = Article
\bibitem[{Bai et~al.(2018)Bai, Sinclair, Tarroni, Oktay, Rajchl, Vaillant, Lee, Aung, Lukaschuk, Sanghvi et~al.}]{bai2018automated}
\bibinfo{author}{W.~Bai}, \bibinfo{author}{M.~Sinclair}, \bibinfo{author}{G.~Tarroni}, \bibinfo{author}{O.~Oktay}, \bibinfo{author}{M.~Rajchl}, \bibinfo{author}{G.~Vaillant}, \bibinfo{author}{A.~M. Lee}, \bibinfo{author}{N.~Aung}, \bibinfo{author}{E.~Lukaschuk}, \bibinfo{author}{M.~M. Sanghvi}, et~al.,
\newblock \bibinfo{title}{Automated cardiovascular magnetic resonance image analysis with fully convolutional networks},
\newblock \bibinfo{journal}{Journal of cardiovascular magnetic resonance} \bibinfo{volume}{20} (\bibinfo{year}{2018}) \bibinfo{pages}{65}.
%Type = Article
\bibitem[{Adams and Bischof(1994)}]{adams1994seeded}
\bibinfo{author}{R.~Adams}, \bibinfo{author}{L.~Bischof},
\newblock \bibinfo{title}{Seeded region growing},
\newblock \bibinfo{journal}{IEEE Transactions on pattern analysis and machine intelligence} \bibinfo{volume}{16} (\bibinfo{year}{1994}) \bibinfo{pages}{641--647}.
%Type = Article
\bibitem[{Chen et~al.(2009)Chen, Smith, Ji, Ward, and Najarian}]{chen2009automated}
\bibinfo{author}{W.~Chen}, \bibinfo{author}{R.~Smith}, \bibinfo{author}{S.-Y. Ji}, \bibinfo{author}{K.~R. Ward}, \bibinfo{author}{K.~Najarian},
\newblock \bibinfo{title}{Automated ventricular systems segmentation in brain ct images by combining low-level segmentation and high-level template matching},
\newblock \bibinfo{journal}{BMC medical informatics and decision making} \bibinfo{volume}{9} (\bibinfo{year}{2009}) \bibinfo{pages}{1--14}.
%Type = Article
\bibitem[{Lang et~al.(2015)Lang, Badano, Mor-Avi, Afilalo, Armstrong, Ernande, Flachskampf, Foster, Goldstein, Kuznetsova et~al.}]{lang2015recommendations}
\bibinfo{author}{R.~M. Lang}, \bibinfo{author}{L.~P. Badano}, \bibinfo{author}{V.~Mor-Avi}, \bibinfo{author}{J.~Afilalo}, \bibinfo{author}{A.~Armstrong}, \bibinfo{author}{L.~Ernande}, \bibinfo{author}{F.~A. Flachskampf}, \bibinfo{author}{E.~Foster}, \bibinfo{author}{S.~A. Goldstein}, \bibinfo{author}{T.~Kuznetsova}, et~al.,
\newblock \bibinfo{title}{Recommendations for cardiac chamber quantification by echocardiography in adults: an update from the american society of echocard and the european association of cardiovascular imaging},
\newblock \bibinfo{journal}{European Heart Journal-Cardiovascular Imaging} \bibinfo{volume}{16} (\bibinfo{year}{2015}) \bibinfo{pages}{233--271}.
%Type = Inproceedings
\bibitem[{Wei et~al.(2020)Wei, Cao, Cao, Zhou, Xue, Ni, and Li}]{wei2020temporal}
\bibinfo{author}{H.~Wei}, \bibinfo{author}{H.~Cao}, \bibinfo{author}{Y.~Cao}, \bibinfo{author}{Y.~Zhou}, \bibinfo{author}{W.~Xue}, \bibinfo{author}{D.~Ni}, \bibinfo{author}{S.~Li},
\newblock \bibinfo{title}{Temporal-consistent segmentation of echocardiography with co-learning from appearance and shape},
\newblock in: \bibinfo{booktitle}{MICCAI 2020: 23rd International Conference, Lima, Peru, October 4--8, 2020, Proceedings, Part II 23}, \bibinfo{organization}{Springer}, pp. \bibinfo{pages}{623--632}.
%Type = Inproceedings
\bibitem[{Myronenko et~al.(2020)Myronenko, Yang, Buch, Xu, Ihsani, Doyle, Michalski, Tenenholtz, and Roth}]{myronenko20204d}
\bibinfo{author}{A.~Myronenko}, \bibinfo{author}{D.~Yang}, \bibinfo{author}{V.~Buch}, \bibinfo{author}{D.~Xu}, \bibinfo{author}{A.~Ihsani}, \bibinfo{author}{S.~Doyle}, \bibinfo{author}{M.~Michalski}, \bibinfo{author}{N.~Tenenholtz}, \bibinfo{author}{H.~Roth},
\newblock \bibinfo{title}{4d cnn for semantic segmentation of cardiac volumetric sequences},
\newblock in: \bibinfo{booktitle}{Statistical Atlases and Computational Models of the Heart. Multi-Sequence CMR Segmentation, CRT-EPiggy and LV Full Quantification Challenges: 10th International Workshop, STACOM 2019, Held in Conjunction with MICCAI 2019, Shenzhen, China, October 13, 2019}, \bibinfo{organization}{Springer}, pp. \bibinfo{pages}{72--80}.
%Type = Inproceedings
\bibitem[{Long et~al.(2015)Long, Shelhamer, and Darrell}]{long2015fully}
\bibinfo{author}{J.~Long}, \bibinfo{author}{E.~Shelhamer}, \bibinfo{author}{T.~Darrell},
\newblock \bibinfo{title}{Fully convolutional networks for semantic segmentation},
\newblock in: \bibinfo{booktitle}{Proceedings of the IEEE conference on CVPR}, pp. \bibinfo{pages}{3431--3440}.
%Type = Inproceedings
\bibitem[{Ronneberger et~al.(2015)Ronneberger, Fischer, and Brox}]{ronneberger2015u}
\bibinfo{author}{O.~Ronneberger}, \bibinfo{author}{P.~Fischer}, \bibinfo{author}{T.~Brox},
\newblock \bibinfo{title}{U-net: Convolutional networks for biomedical image segmentation},
\newblock in: \bibinfo{booktitle}{MICCAI 2015: 18th international conference, Munich, Germany, October 5-9, 2015, proceedings, part III 18}, \bibinfo{organization}{Springer}, pp. \bibinfo{pages}{234--241}.
%Type = Article
\bibitem[{Litjens et~al.(2017)Litjens, Kooi, Bejnordi, Setio, Ciompi, Ghafoorian, Van Der~Laak, Van~Ginneken, and S{\'a}nchez}]{litjens2017survey}
\bibinfo{author}{G.~Litjens}, \bibinfo{author}{T.~Kooi}, \bibinfo{author}{B.~E. Bejnordi}, \bibinfo{author}{A.~A.~A. Setio}, \bibinfo{author}{F.~Ciompi}, \bibinfo{author}{M.~Ghafoorian}, \bibinfo{author}{J.~A. Van Der~Laak}, \bibinfo{author}{B.~Van~Ginneken}, \bibinfo{author}{C.~I. S{\'a}nchez},
\newblock \bibinfo{title}{A survey on deep learning in medical image analysis},
\newblock \bibinfo{journal}{Medical image analysis} \bibinfo{volume}{42} (\bibinfo{year}{2017}) \bibinfo{pages}{60--88}.
%Type = Inproceedings
\bibitem[{Hatamizadeh~et al.(2022)}]{hatamizadeh2022unetr}
\bibinfo{author}{A.~Hatamizadeh~et al.},
\newblock \bibinfo{title}{Unetr: Transformers for 3d medical image segmentation},
\newblock in: \bibinfo{booktitle}{Proceedings of the IEEE/CVF winter conference on applications of computer vision}, pp. \bibinfo{pages}{574--584}.
%Type = Inproceedings
\bibitem[{Hatamizadeh et~al.(2021)}]{hatamizadeh2021swin}
\bibinfo{author}{A.~Hatamizadeh}, et~al.,
\newblock \bibinfo{title}{Swin unetr: Swin transformers for semantic segmentation of brain tumors in mri images},
\newblock in: \bibinfo{booktitle}{International MICCAI Brainlesion Workshop}, \bibinfo{organization}{Springer}, pp. \bibinfo{pages}{272--284}.
%Type = Article
\bibitem[{Chan et~al.(2021)Chan, Zheng, Wiputra, Leo, and Yap}]{chan2021full}
\bibinfo{author}{W.~X. Chan}, \bibinfo{author}{Y.~Zheng}, \bibinfo{author}{H.~Wiputra}, \bibinfo{author}{H.~L. Leo}, \bibinfo{author}{C.~H. Yap},
\newblock \bibinfo{title}{Full cardiac cycle asynchronous temporal compounding of 3d echocardiography images},
\newblock \bibinfo{journal}{Medical Image Analysis} \bibinfo{volume}{74} (\bibinfo{year}{2021}) \bibinfo{pages}{102229}.
%Type = Article
\bibitem[{Xue et~al.(2022)Xue, Cao, Ma, Bai, Wang, and Ni}]{xue2022improved}
\bibinfo{author}{W.~Xue}, \bibinfo{author}{H.~Cao}, \bibinfo{author}{J.~Ma}, \bibinfo{author}{T.~Bai}, \bibinfo{author}{T.~Wang}, \bibinfo{author}{D.~Ni},
\newblock \bibinfo{title}{Improved segmentation of echocardiography with orientation-congruency of optical flow and motion-enhanced segmentation},
\newblock \bibinfo{journal}{IEEE Journal of Biomedical and Health Informatics} \bibinfo{volume}{26} (\bibinfo{year}{2022}) \bibinfo{pages}{6105--6115}.
%Type = Inproceedings
\bibitem[{Li et~al.(2019)Li, Zhang, Yang, Wang, Zhang, Liu, Zheng, and Li}]{li2019recurrent}
\bibinfo{author}{M.~Li}, \bibinfo{author}{W.~Zhang}, \bibinfo{author}{G.~Yang}, \bibinfo{author}{C.~Wang}, \bibinfo{author}{H.~Zhang}, \bibinfo{author}{H.~Liu}, \bibinfo{author}{W.~Zheng}, \bibinfo{author}{S.~Li},
\newblock \bibinfo{title}{Recurrent aggregation learning for multi-view echocardiographic sequences segmentation},
\newblock in: \bibinfo{booktitle}{MICCAI 2019: 22nd International Conference, Shenzhen, China, October 13--17, 2019, Proceedings, Part II 22}, \bibinfo{organization}{Springer}, pp. \bibinfo{pages}{678--686}.
%Type = Article
\bibitem[{Wei et~al.(2023)Wei, Ma, Zhou, Xue, and Ni}]{wei2023co}
\bibinfo{author}{H.~Wei}, \bibinfo{author}{J.~Ma}, \bibinfo{author}{Y.~Zhou}, \bibinfo{author}{W.~Xue}, \bibinfo{author}{D.~Ni},
\newblock \bibinfo{title}{Co-learning of appearance and shape for precise ejection fraction estimation from echocardiographic sequences},
\newblock \bibinfo{journal}{Medical Image Analysis} \bibinfo{volume}{84} (\bibinfo{year}{2023}) \bibinfo{pages}{102686}.
%Type = Article
\bibitem[{Wu et~al.(2022)Wu, Liu, Xiao, Wen, Cheng, and Qin}]{wu2022semi}
\bibinfo{author}{H.~Wu}, \bibinfo{author}{J.~Liu}, \bibinfo{author}{F.~Xiao}, \bibinfo{author}{Z.~Wen}, \bibinfo{author}{L.~Cheng}, \bibinfo{author}{J.~Qin},
\newblock \bibinfo{title}{Semi-supervised segmentation of echocardiography videos via noise-resilient spatiotemporal semantic calibration and fusion},
\newblock \bibinfo{journal}{Medical Image Analysis} \bibinfo{volume}{78} (\bibinfo{year}{2022}) \bibinfo{pages}{102397}.
%Type = Inproceedings
\bibitem[{Ahn et~al.(2021)Ahn, Ta, Thorn, Langdon, Sinusas, and Duncan}]{ahn2021multi}
\bibinfo{author}{S.~S. Ahn}, \bibinfo{author}{K.~Ta}, \bibinfo{author}{S.~Thorn}, \bibinfo{author}{J.~Langdon}, \bibinfo{author}{A.~J. Sinusas}, \bibinfo{author}{J.~S. Duncan},
\newblock \bibinfo{title}{Multi-frame attention network for left ventricle segmentation in 3d echocardiography},
\newblock in: \bibinfo{booktitle}{MICCAI 2021: 24th International Conference, Strasbourg, France, September 27--October 1, 2021, Proceedings, Part I 24}, \bibinfo{organization}{Springer}, pp. \bibinfo{pages}{348--357}.
%Type = Article
\bibitem[{Dai et~al.(2024)Dai, Dong, Yan, Sun, Zhang, Li, and Xu}]{dai2024i2u}
\bibinfo{author}{D.~Dai}, \bibinfo{author}{C.~Dong}, \bibinfo{author}{Q.~Yan}, \bibinfo{author}{Y.~Sun}, \bibinfo{author}{C.~Zhang}, \bibinfo{author}{Z.~Li}, \bibinfo{author}{S.~Xu},
\newblock \bibinfo{title}{I2u-net: A dual-path u-net with rich information interaction for medical image segmentation},
\newblock \bibinfo{journal}{Medical Image Analysis}  (\bibinfo{year}{2024}) \bibinfo{pages}{103241}.
%Type = Article
\bibitem[{Cai et~al.(2024)Cai, Lu, Wu, Berretti, and Wan}]{cai2024dt}
\bibinfo{author}{Y.~Cai}, \bibinfo{author}{H.~Lu}, \bibinfo{author}{S.~Wu}, \bibinfo{author}{S.~Berretti}, \bibinfo{author}{S.~Wan},
\newblock \bibinfo{title}{Dt-vnet: Deep transformer-based vnet framework for 3d prostate mri segmentation},
\newblock \bibinfo{journal}{IEEE Journal of Biomedical and Health Informatics}  (\bibinfo{year}{2024}).
%Type = Article
\bibitem[{Leclerc et~al.(2019)Leclerc, Smistad, Pedrosa, {\O}stvik, Cervenansky, Espinosa, Espeland, Berg, Jodoin, Grenier et~al.}]{leclerc2019deep}
\bibinfo{author}{S.~Leclerc}, \bibinfo{author}{E.~Smistad}, \bibinfo{author}{J.~Pedrosa}, \bibinfo{author}{A.~{\O}stvik}, \bibinfo{author}{F.~Cervenansky}, \bibinfo{author}{F.~Espinosa}, \bibinfo{author}{T.~Espeland}, \bibinfo{author}{E.~A.~R. Berg}, \bibinfo{author}{P.-M. Jodoin}, \bibinfo{author}{T.~Grenier}, et~al.,
\newblock \bibinfo{title}{Deep learning for segmentation using an open large-scale dataset in 2d echocardiography},
\newblock \bibinfo{journal}{IEEE transactions on medical imaging} \bibinfo{volume}{38} (\bibinfo{year}{2019}) \bibinfo{pages}{2198--2210}.
%Type = Article
\bibitem[{Zhao et~al.(2023)Zhao, Ferdian, Maso~Talou, Quill, Gilbert, Wang, Babarenda~Gamage, Pedrosa, D’hooge, Sutton et~al.}]{zhao2023mitea}
\bibinfo{author}{D.~Zhao}, \bibinfo{author}{E.~Ferdian}, \bibinfo{author}{G.~D. Maso~Talou}, \bibinfo{author}{G.~M. Quill}, \bibinfo{author}{K.~Gilbert}, \bibinfo{author}{V.~Y. Wang}, \bibinfo{author}{T.~P. Babarenda~Gamage}, \bibinfo{author}{J.~Pedrosa}, \bibinfo{author}{J.~D’hooge}, \bibinfo{author}{T.~M. Sutton}, et~al.,
\newblock \bibinfo{title}{Mitea: A dataset for machine learning segmentation of the left ventricle in 3d echocardiography using subject-specific labels from cardiac magnetic resonance imaging},
\newblock \bibinfo{journal}{Frontiers in Cardiovascular Medicine} \bibinfo{volume}{9} (\bibinfo{year}{2023}) \bibinfo{pages}{1016703}.
%Type = Article
\bibitem[{Bernard et~al.(2018)Bernard, Lalande, Zotti, Cervenansky, Yang, Heng, Cetin, Lekadir, Camara, Ballester et~al.}]{bernard2018deep}
\bibinfo{author}{O.~Bernard}, \bibinfo{author}{A.~Lalande}, \bibinfo{author}{C.~Zotti}, \bibinfo{author}{F.~Cervenansky}, \bibinfo{author}{X.~Yang}, \bibinfo{author}{P.-A. Heng}, \bibinfo{author}{I.~Cetin}, \bibinfo{author}{K.~Lekadir}, \bibinfo{author}{O.~Camara}, \bibinfo{author}{M.~A.~G. Ballester}, et~al.,
\newblock \bibinfo{title}{Deep learning techniques for automatic mri cardiac multi-structures segmentation and diagnosis: is the problem solved?},
\newblock \bibinfo{journal}{IEEE transactions on medical imaging} \bibinfo{volume}{37} (\bibinfo{year}{2018}) \bibinfo{pages}{2514--2525}.
%Type = Inproceedings
\bibitem[{Lu et~al.(2019)Lu, Wang, Ma, Shen, Shao, and Porikli}]{lu2019see}
\bibinfo{author}{X.~Lu}, \bibinfo{author}{W.~Wang}, \bibinfo{author}{C.~Ma}, \bibinfo{author}{J.~Shen}, \bibinfo{author}{L.~Shao}, \bibinfo{author}{F.~Porikli},
\newblock \bibinfo{title}{See more, know more: Unsupervised video object segmentation with co-attention siamese networks},
\newblock in: \bibinfo{booktitle}{Proceedings of the IEEE/CVF conference on computer vision and pattern recognition}, pp. \bibinfo{pages}{3623--3632}.
%Type = Article
\bibitem[{Shaker et~al.(2024)Shaker, Maaz, Rasheed, Khan, Yang, and Khan}]{shaker2024unetr++}
\bibinfo{author}{A.~M. Shaker}, \bibinfo{author}{M.~Maaz}, \bibinfo{author}{H.~Rasheed}, \bibinfo{author}{S.~Khan}, \bibinfo{author}{M.-H. Yang}, \bibinfo{author}{F.~S. Khan},
\newblock \bibinfo{title}{Unetr++: delving into efficient and accurate 3d medical image segmentation},
\newblock \bibinfo{journal}{IEEE Transactions on Medical Imaging} \bibinfo{volume}{43} (\bibinfo{year}{2024}) \bibinfo{pages}{3377--3390}.

\end{thebibliography}

\end{document}